\documentclass{article}
\PassOptionsToPackage{numbers, compress, sort}{natbib}

\usepackage[table, dvipsnames]{xcolor}
\usepackage[final]{corl_2026} %
\hypersetup{pdftitle={AmodalBox: Automated 3D Bounding Box Annotation under Partial Visibility},pdfauthor={In-Jae Lee, Namtae Kim, Robert Exley, Kwonyoung Ryu, Jaesik Park}}
\usepackage[utf8]{inputenc} %
\usepackage[T1]{fontenc}    %
\usepackage{hyperref}       %
\usepackage{url}            %
\usepackage{booktabs}       %
\usepackage{amsfonts}       %
\usepackage{nicefrac}       %
\usepackage{microtype}      %
\usepackage{makecell}%
\definecolor{my_lime}{rgb}{0,0.8,0}
\definecolor{mycyan}{rgb}{0,1,1}
\definecolor{myred}{rgb}{1,0,0}
\usepackage{algorithm}
\usepackage{graphicx}
\usepackage{algpseudocode}
\usepackage{amssymb}
\usepackage{pifont}
\usepackage{amsmath} 
\usepackage{multirow}
\usepackage{booktabs}
\usepackage{subcaption}
\usepackage{wrapfig}
\usepackage{tabularx}
\newcommand{\padnum}[1]{\phantom{0}#1}

\usepackage[nameinlink]{cleveref} %
\Crefname{section}{Sec.}{Secs.}   %
\Crefname{figure}{Fig.}{Figs.}
\Crefname{table}{Table}{Tables}
\crefname{equation}{}{}           %
\Crefname{equation}{}{}
\newcommand{\Ours}{AmodalBox}
\title{\Ours: Automated 3D Bounding Box Annotation under Partial Visibility}

\author{%
  In-Jae Lee\textsuperscript{1} \quad
  Namtae Kim\textsuperscript{1} \quad
  Robert Exley\textsuperscript{2} \quad
  Kwonyoung Ryu\textsuperscript{3} \quad
  Jaesik Park\textsuperscript{1} \\
  \textsuperscript{1}Seoul National University \quad
  \textsuperscript{2}University of Pittsburgh \quad
  \textsuperscript{3}POSTECH
}

\begin{document}
\maketitle

\vspace{-30pt}
\begin{figure}[!h]
    \centering
    \includegraphics[width=1.0\linewidth]{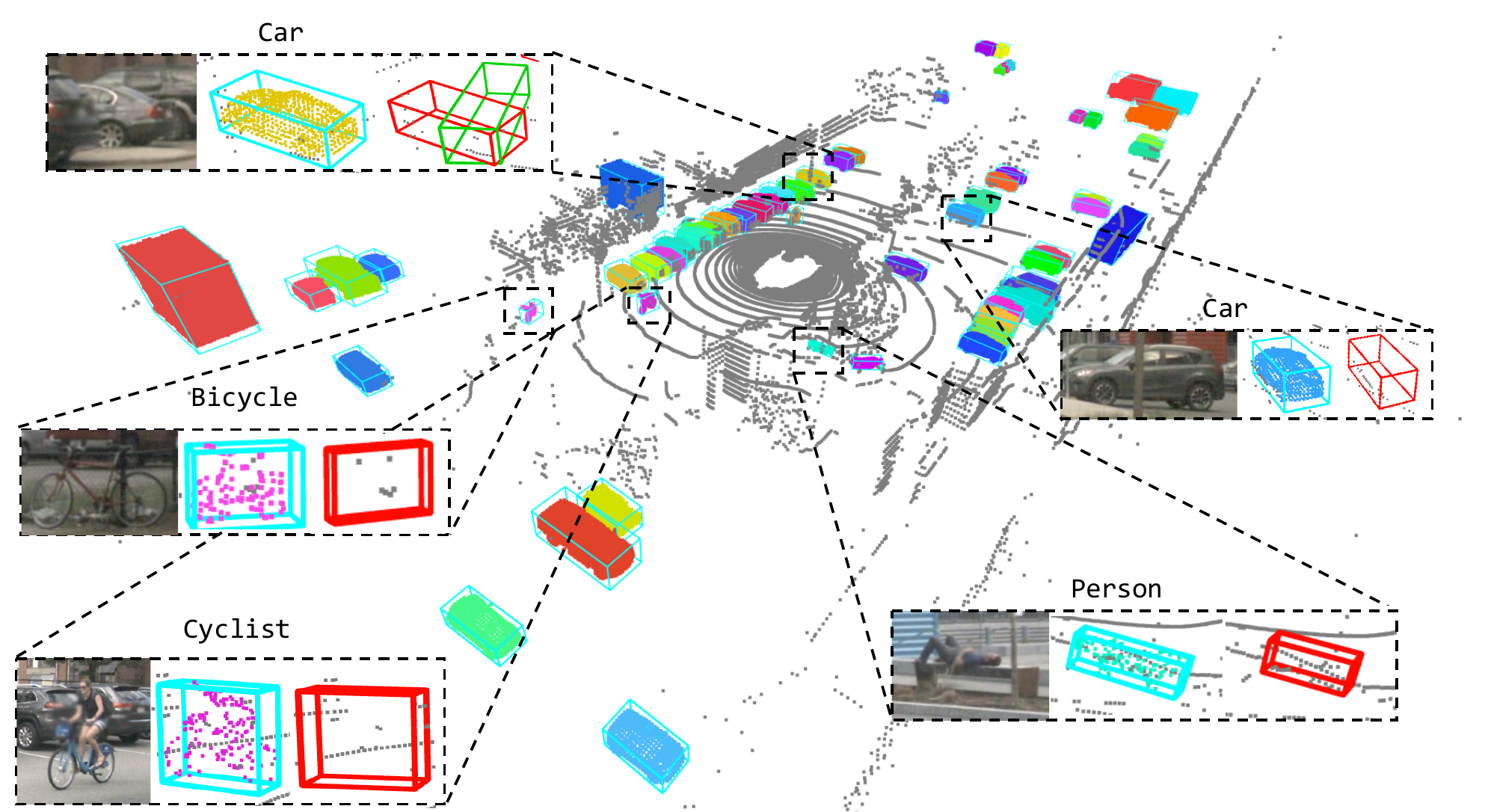}
    \caption{\textbf{Annotation from \Ours{}.}
    \label{fig:figure1}
For diverse classes in the nuScenes dataset~\cite{nuscenes}, \textcolor{my_lime}{Baseline} often struggles under partial visibility, either missing objects or producing mislocalized boxes. 
In contrast, \textcolor{mycyan}{AmodalBox} captures more complete object extents and produces spatially coherent annotations, while \textcolor{myred}{human annotations} are shown as reference boxes. Best viewed in color and zoomed in.}
\end{figure}

\begin{abstract}
Recent progress in vision foundation models (VFMs) has made automated open-vocabulary 3D bounding box annotation increasingly feasible by jointly leveraging LiDAR point clouds and images. However, partial observations remain a major challenge in sparse driving scenes, where visible LiDAR points often cover only a small fraction of the object's surface and provide insufficient cues for estimating its full spatial extent. As a result, existing methods often produce mislocalized boxes or leave objects unlabeled. To address this limitation, we propose \textbf{\Ours{}}, an automated annotation framework that leverages image-generated 3D assets as amodal geometric priors for partially observed objects. By establishing correspondences between input images and rendered views of 3D assets, \Ours{} robustly aligns canonical amodal geometry with partially observed world-space LiDAR point clouds. To improve temporal consistency and reduce computational cost, \Ours{} 
generates a single asset per instance, estimates its metric scale once at a single frame and propagates it across frames for dynamic instances, and aggregates multi-frame observations for static instances.
3D object detectors trained on annotations from \Ours{} improve performance over strong baselines by 14.1 mAP on the nuScenes dataset and by 12.45 AP at IoU 0.5 for the vehicle class on the Waymo Open Dataset (WOD). Our project page is available at: \href{https://oliver0922.github.io/AmodalBox/}{https://oliver0922.github.io/AmodalBox/}
\end{abstract}

\keywords{Automated Annotation, 3D Object Detection, Multi-modal Fusion, Autonomous Driving} 
\section{Introduction}
\label{sec:introduction}

3D object detection is a fundamental capability for robotics and autonomous driving. While closed-set 3D detectors~\cite{bevfusion,crab,crn,bevformer} have achieved strong performance on predefined categories, real-world scenes contain previously unseen objects that may still affect downstream prediction and planning. This motivates open-vocabulary 3D perception~\cite{ovscan,findnpropagate,opensight,upvl}, which aims to recognize and localize object categories beyond a fixed label space~\cite{openbox}.

Despite this progress, scaling open-vocabulary 3D detection remains fundamentally limited by the lack of large-scale 3D annotations for diverse object categories. Existing 3D detectors require instance-level bounding boxes that accurately specify object category, position, orientation, and spatial extent. However, obtaining such annotations is expensive and labor-intensive~\cite{detectany3d,wilddet3d}. This bottleneck becomes even more severe at scale, where annotations must cover a much wider diversity of object shapes, sizes, and appearances. Therefore, automated annotation provides a practical path toward scaling open-vocabulary 3D perception.

To alleviate the annotation bottleneck, earlier unsupervised methods~\cite{cpd,oyster,lise} refine noisy initial pseudo-labels through heuristic scoring and bounding box filtering~\cite{cpd, modest,findnpropagate,ovscan} or computationally expensive self-training~\cite{modest,lise}. More recent methods~\cite{union,ovscan,openbox} leverage VFMs~\cite{samv2,sam,groundingdino,owlvit,dinov2} to provide semantic information to LiDAR point clouds.
However, partial observations remain difficult to handle, since visible evidence alone provides only incomplete constraints on the object's full spatial extent. Existing methods mitigate this issue through temporal refinement, size priors, or noisy point cloud filtering, but these strategies mainly refine boxes within the space supported by observed points. As a result, they may still produce mislocalized boxes or false negatives, as illustrated in~\Cref{fig:figure1}. These noisy annotations can in turn degrade 3D object detector training, leading to weaker supervision for occluded objects. This motivates us to propose an annotation pipeline that uses visible observations as alignment cues while leveraging amodal object geometry to infer the full object extent beyond what is directly observed.

In this paper, we propose \textbf{\Ours{}}, a novel automated 3D bounding box annotation pipeline for handling partial visibility in real-world driving scenes. Our method aligns partial world-space LiDAR observations with an amodal 3D asset generated in canonical space by an image-to-3D generation model~\cite{fastsam3d}. To guide this 3D alignment, we leverage correspondences obtained from 2D matching~\cite{roma} between the real image and rendered views of the generated asset. \Ours{} further accounts for instance motion during alignment. For dynamic instances, our method estimates the alignment once at a single anchor frame and propagates it across the sequence, while for static instances, it aggregates LiDAR point clouds from multiple frames before alignment. As a result, \Ours{} improves 3D bounding box localization accuracy and yields more robust alignment under occlusion. Through experiments on the nuScenes~\cite{nuscenes} dataset and WOD~\cite{waymo}, we show that leveraging image-generated amodal geometry yields more reliable 3D annotations and consistently improves downstream 3D object detection performance.

Our contributions are:
\begin{itemize}
    \item We propose \Ours{}, an open-vocabulary automated 3D bounding box annotation pipeline for driving scenes that explicitly addresses partial visibility and occlusion.
    \item To tackle partial visibility, we introduce an amodal 3D alignment strategy that leverages image-generated object geometry and correspondences between real images and rendered views of the generated 3D asset to align partial world-space LiDAR observations with canonical 3D assets.
    \item We further incorporate motion-aware temporal reasoning, using multi-frame aggregation for static instances and anchor-frame propagation for moving instances to improve annotation consistency across frames.
    \item Training 3D object detectors on annotations from \Ours{} outperforms strong automated annotation baselines by 14.1 mAP on the nuScenes dataset~\cite{nuscenes} and by 12.45 AP for the vehicle class at IoU 0.5 on WOD~\cite{waymo}.
\end{itemize}

\section{Related Work}
\label{sec:related}
\subsection{Open-vocabulary 3D Object Detection}
OpenSight~\cite{opensight} and Find n' Propagate~\cite{findnpropagate} lift 2D proposals from open-vocabulary 2D models into 3D space to generate pseudo 3D boxes, and further improve them through geometric search, alignment, or propagation. OVM3D-Det~\cite{ovm3ddet} performs such lifting from a single image, fitting boxes to pseudo-
LiDAR obtained from a monocular depth model with category size priors to resolve the scale
ambiguity. UP-VL~\cite{upvl} extends motion-based auto-labeling~\cite{miup} with vision-language cues, producing semantically aligned amodal boxes and distilling point-wise semantic features into a 3D detector. OV-SCAN~\cite{ovscan} improves novel object discovery and cross-modal alignment by optimizing 3D box search, filtering semantically inconsistent 3D--2D pairs, and applying a hierarchical alignment head. AnnofreeOD~\cite{annofreeod} models per-face box uncertainty and learns from noisy pseudo labels through box augmentation and noise-aware regression. FOMO-3D~\cite{fomo3d} refines 3D boxes from LiDAR and camera branches in BEV space using attention. Finally, OpenBox~\cite{openbox} performs cross-modal instance alignment and generates adaptive boxes according to the physical type of each instance. 
\subsection{Unsupervised 3D Object Detection}

Prior work on unsupervised 3D object detection typically bootstraps pseudo labels from LiDAR geometry, motion cues, or temporal consistency. LiDAR-centric methods such as MODEST~\cite{modest}, OYSTER~\cite{oyster}, and CPD~\cite{cpd} rely on clustering, temporal refinement, or prototype-based regularization to improve noisy boxes. Multi-modal extensions further incorporate image cues. MI-UP~\cite{miup}, LiSe~\cite{lise}, and UNION~\cite{union} combine motion, appearance, or cross-modal cues to improve pseudo-label quality, while CM3D~\cite{cm3d} exploits image-based foundation models to generate pseudo 3D boxes for pre-training. PLOT~\cite{plot} brings temporal aggregation to monocular video, tracking object points
across frames and registering them into pseudo-LiDAR for both static and dynamic objects. Despite these advances, most existing methods still derive supervision primarily from visible evidence in images and sparse point clouds, and then refine it through clustering, propagation, or regularization. In contrast, our method introduces an explicit amodal shape prior from image-to-3D generation, making the annotation process less dependent on incomplete observed LiDAR geometry.
\vspace{-0.2em}
\subsection{Automated Annotation with 3D Shape Priors}

Prior works generate object-level 3D annotations by fitting predefined shape priors to visual and geometric observations.
CAD-based methods align retrieved object models with RGB-D scans or RGB videos to obtain object-level shape and pose annotations~\cite{scan2cad,cad_rgbd,cadestate}, while template-based methods exploit category-specific deformable models for shape-aware annotation~\cite{autoshape}.
Other methods use implicit or canonical shape priors to recover object pose, shape, and 3D extent, where NOCS~\cite{nocs}-based methods rely on canonical coordinate prediction~\cite{neurocs,omninocs}, while SDF-based methods optimize object geometry through differentiable rendering~\cite{sdflabel,vsrd}.
Segment, Lift and Fit~\cite{slf} further uses SAM masks from 2D prompts and lifts them into 3D through SDF-based shape fitting with LiDAR and mask projection consistency.
Although these methods reduce the need for manual 3D annotations, they rely on predefined CAD repositories, category-specific templates, or SDF/NOCS shape spaces.
More recently, LabelAny3D~\cite{labelany3d} replaces such predefined priors with a generated mesh, whose pose is recovered by PnP from image-to-render matches and whose metric scale is obtained from monocular depth.
\Ours{} also uses a generated asset~\cite{sam3d,fastsam3d} as the shape prior, but lifts the matches to 3D-3D correspondences with LiDAR points, solves scale, yaw, and translation jointly in metric space, and propagates the alignment over time.

\begin{figure}[!t]
    \centering
    \includegraphics[width=\linewidth]{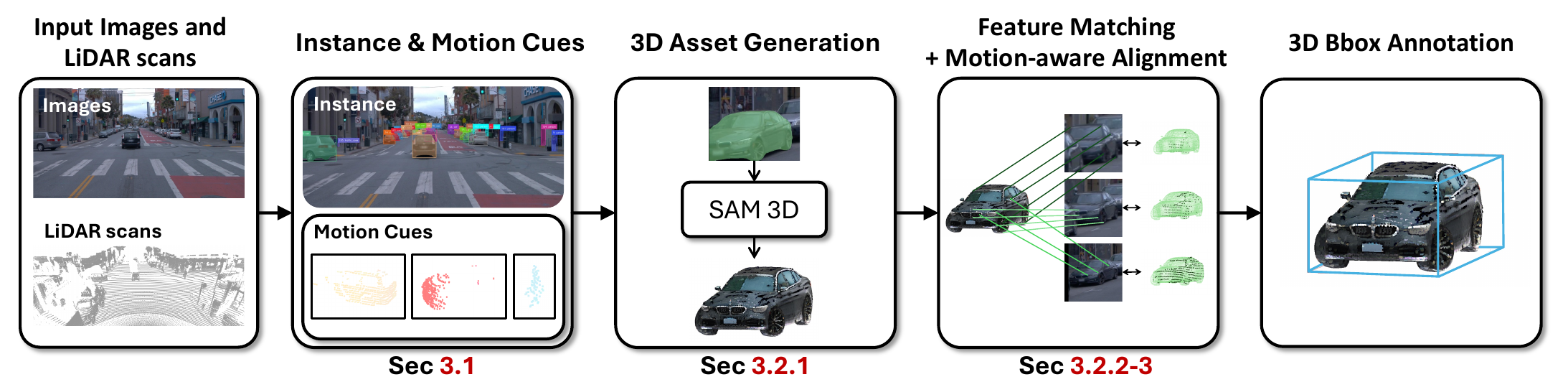}
  \caption{\textbf{Overview of \Ours{}}. Given time-synchronized images and LiDAR scans, we first extract instance-level cues and estimate the motion state of each instance using VFMs~\cite{groundingdino,samv2}, scene flow~\cite{deltaflow}, and PP scores~\cite{modest}. We then align the amodal 3D prior~\cite{fastsam3d} with the partial LiDAR point cloud across their coordinate spaces using correspondences obtained from image-level matching. Finally, we generate a 3D bounding box annotation from the aligned amodal asset.}
    \label{fig:fig2}
\end{figure}

\section{AmodalBox}
\label{sec:method}
We propose \Ours{}, an automated 3D bounding box annotation framework for partially visible objects, as illustrated in Fig.~\ref{fig:fig2}. \Ours{} aligns partial world-space LiDAR observations with a canonical amodal 3D asset generated from images. To bridge the two spaces, we use feature matching between real images and rendered asset views, and perform motion-aware alignment for static and dynamic instances.
We first introduce the preliminaries in Sec.~\ref{method/sub/3.1}. Sec.~\ref{method/sub/3.2} then presents our cross-modal spatial alignment strategy.
\subsection{Preliminaries}
\label{method/sub/3.1}

\noindent\textbf{Instance, Shape, and Motion Cues} 
Following~\cite{openbox}, \Ours{} considers both instance identity and motion state. To extract instance-level cues, \Ours{} processes each input image $I_t^n$ with an open-vocabulary 2D detector $f$~\cite{groundingdino} and a trackable open-vocabulary instance segmentation model $h$~\cite{samv2}, using a set of text queries $\mathcal{T}$. We denote the resulting 2D instance cues at frame $t$ and view $n$ by
\begin{equation}
\mathcal{S}_t^n = h(f(I_t^n, \mathcal{T})),
\end{equation}
where $\mathcal{S}_t^n$ includes the 2D boxes $B_t^n$, semantic labels $C_t^n$, instance masks $M_t^n$, and temporally consistent instance IDs for the corresponding view. 
Using the calibration matrix between the camera and LiDAR, we gather LiDAR points within each mask to obtain a set of instance-level point clouds, defined as
\begin{equation}
\mathbb{P}_t^n = \{\mathcal{P}_t^{n,k}\}_{k=1}^{K_t^n}, 
\qquad
\mathcal{P}_t^{n,k} \in \mathbb{R}^{M_t^{n,k} \times 3},
\end{equation}
where $K_t^n$ denotes the number of instances in view $n$ at frame $t$, and $M_t^{n,k}$ denotes the number of LiDAR points associated with the $k$-th instance. Instances observed from multiple cameras are associated through shared LiDAR points and class-wise BEV NMS (Appendix~\ref{appendix:cross-view}); hereafter we omit the view index $n$. 

In addition, as described in Sec.~\ref{method/sub/3.2}, we leverage scene flow~\cite{deltaflow} on nuScenes~\cite{nuscenes} and the persistence point score (PP score)~\cite{modest}, which measures point-level ephemerality, on WOD~\cite{waymo}, respectively, to improve alignment accuracy. For each instance-level point cloud, we compute point-wise motion cues using the corresponding signal and assign its motion state by majority voting over the associated points.

\subsection{Cross-Modal Spatial Alignment}
\label{method/sub/3.2}
\noindent\textbf{3D Asset Generation} Partial visibility makes it difficult to recover the full spatial extent of an object from visible observations alone, often resulting in mislocalized 3D bounding boxes. To overcome this limitation, we leverage 3D Gaussians~\cite{3dgs} as an amodal 3D asset using an image-to-3D generation model $\psi$~\cite{fastsam3d}. In autonomous driving datasets, however, a scene often contains many object instances, and repeatedly using all image observations of the same instance across different frames is computationally expensive and can also produce inconsistent shapes. Therefore, for each instance $k$, we select the anchor image frame $I_{anchor,k}$ in which its mask has the largest area and feed the corresponding image-mask pair into $\psi$. This yields an amodal 3D shape in canonical space, represented as a set of 3D Gaussians defined as:
\begin{equation}
\mathcal{G}^{k} = \psi(I_{anchor,k}, M_{anchor,k}).
\end{equation}
We then extract the Gaussian centers from $\mathcal{G}^{k}$ and use them as the amodal point representation, denoted by $A^{k}=\{\mu_j\}_{j=1}^{J}$, where $\mu_j \in \mathbb{R}^3$ is the center of the $j$-th Gaussian. Since $A^{k}$ is defined in canonical space, we align it with the partial LiDAR point cloud $\mathcal{P}_{anchor}^{k}$ in world space, which captures only the visible geometry of the object. The problem is therefore formulated as estimating a relative Sim(3) transformation, including scale, rotation, and translation, that best aligns the two representations across modalities.

\begin{figure}[!t]
    \centering
    \includegraphics[width=\linewidth]{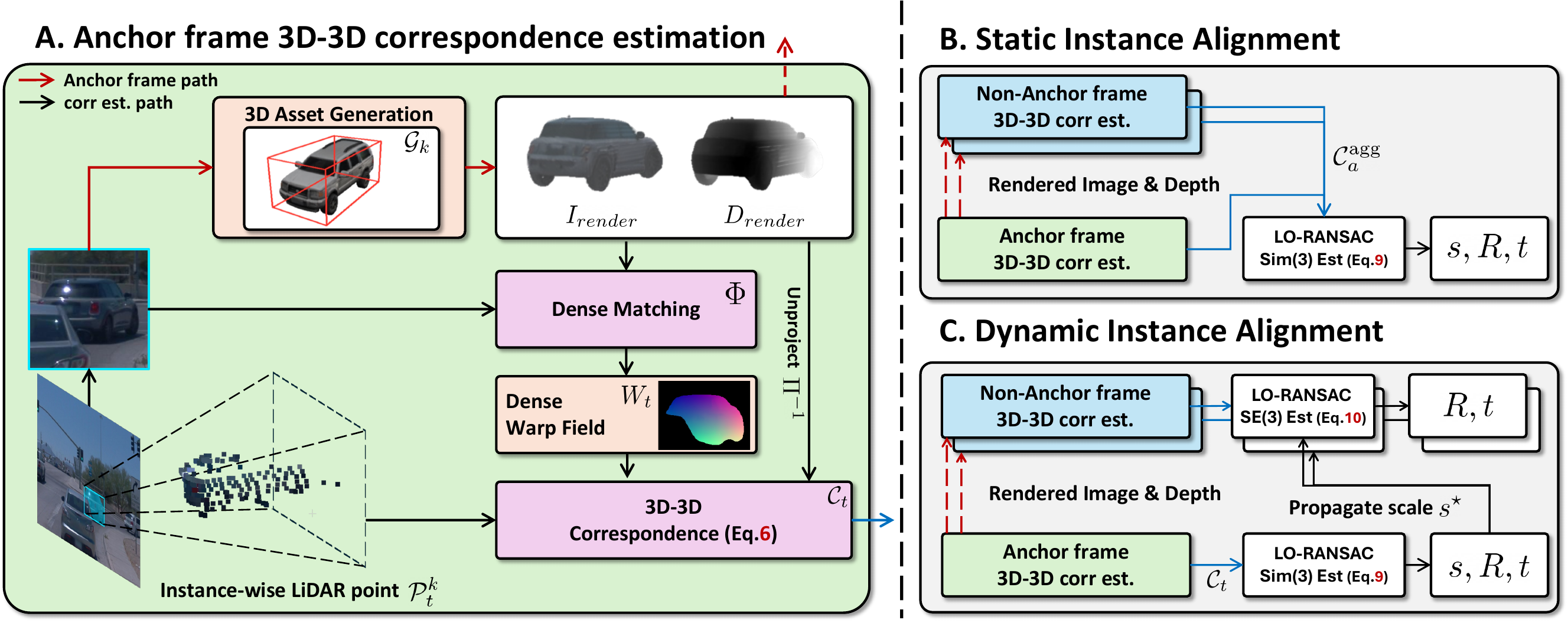}
    \caption{\textbf{Motion-aware Alignment.}
(A) We generate a 3D asset from the anchor-frame image using~\cite{fastsam3d}, render $I_{\mathrm{render}}$ and $D_{\mathrm{render}}$, and obtain 3D--3D correspondences via image matching and depth unprojection.
(B) Using the anchor-rendered $I_{\mathrm{render}}$ and $D_{\mathrm{render}}$, we aggregate neighboring-frame correspondences and estimate the alignment (Eq.~\cref{eq:eq9}).
(C) We estimate the anchor scale $s^\star$ (Eq.~\cref{eq:eq9}), propagate it across frames, and align each frame (Eq.~\cref{eq:eq10}).}
\end{figure}

\noindent\textbf{Feature Matching} Directly aligning these partial observations with the canonical-space amodal shape $A^{k}$ using only geometric registration, such as ICP~\cite{icp}, is prone to fall in local minima when the initialization is inaccurate, often resulting in mislocalized bounding boxes. %
Instead, we use image-space matching as a robust cue. The 3D Gaussian asset is rendered from candidate views~\cite{3dgs}, and we select the rendered view used for matching according to the view-selection procedure described in Appendix~\ref{appendix:view-selection}. Visual correspondences between the selected rendered view and the real image provide alignment cues even when the LiDAR observation is sparse.
We perform dense matching $\Phi$~\cite{roma} between the rendered image $I_{\mathrm{render}}$ and the corresponding real image $I_t$, producing a dense warp field from real-image pixels $u^{\mathrm{real}}_{t}$ to rendered-image pixels $u^{\mathrm{render}}_{t}$ as
\begin{equation}
W_t = \Phi\!\left(I_t, I_{\mathrm{render}}\right),
\qquad
u^{\mathrm{render}}_{t} = W_t(u^{\mathrm{real}}_{t}).
\end{equation}
Although $W_t$ is dense, we do not lift all image pixels into 3D. Instead, the real-side matching candidates are restricted to projected LiDAR pixels. Specifically, each instance-level LiDAR point $\mathbf{x}_j\in\mathcal{P}_t^k$ is projected onto the real image as
\begin{equation}
u_{t,j}^{\mathrm{real}}
=
\pi_t(\mathbf{x}_j),
\end{equation}
where $\pi_t(\cdot)$ is the LiDAR-to-image projection. We then sample the dense warp map only at these projected LiDAR pixels and sample the corresponding rendered depth as
\begin{equation}
u_{t,j}^{\mathrm{render}}
= W_t\!\left(u_{t,j}^{\mathrm{real}}\right),\;
d_{t,j}^{\mathrm{render}}
= D_{\mathrm{render}}\!\left(u_{t,j}^{\mathrm{render}}\right),\;
\mathbf{p}_{t,j}^{\mathrm{amodal}}
= \Pi_{\mathrm{render}}^{-1}
\left(u_{t,j}^{\mathrm{render}}, d_{t,j}^{\mathrm{render}}\right).
\end{equation}
Thus, the partial 3D point is the original LiDAR point $\mathbf{x}_j$, while the amodal 3D point is obtained by unprojecting the matched rendered pixel with its rendered depth. After discarding invalid projections, warp samples, and rendered depths, we obtain the lifted 3D--3D correspondence set
$\mathcal{C}_t=\{(\mathbf{p}^{\mathrm{amodal}}_{t,j},\mathbf{p}^{\mathrm{partial}}_{t,j})\}_{j=1}^{L_t}$,
where $\mathbf{p}^{\mathrm{partial}}_{t,j}$ denotes the LiDAR point $\mathbf{x}_j$ of the $j$-th surviving correspondence.

\noindent\textbf{Motion-aware Alignment} Given the lifted correspondences $\mathcal{C}_t$, we estimate a transformation from the canonical amodal asset to the world-space partial observation of the target instance. 
In autonomous driving scenes, most object categories can be reasonably approximated by a single yaw angle, and their coarse physical size can be represented by category-level size priors following OpenBox~\cite{openbox}. 
Based on this, we formulate the alignment as a scale-constrained 5-DoF problem, consisting of scale $s>0$, translation $\mathbf{t} \in \mathbb{R}^3$, and yaw rotation $R_{\mathrm{yaw}}\in SO(3)$, so that each correspondence satisfies
\begin{equation}
\mathbf{p}^{\mathrm{partial}}_{t,j} \approx s \, R_{\mathrm{yaw}}(\theta)\, \mathbf{p}^{\mathrm{amodal}}_{t,j} + \mathbf{t}.
\end{equation}
We estimate this transformation robustly using LO-RANSAC~\cite{loransac}, where a closed-form yaw-constrained Umeyama-style estimator is used as the minimal solver for hypothesis generation~\cite{umeyama}. Details on the size prior and the solver are given in Appendix~\ref{appendix:motion-aware}.

\Ours{} treats static and dynamic instances differently, since the former benefit from multi-frame aggregation while the latter require temporally stable propagation from a single anchor alignment.
For static instances, we match the anchor-rendered view $I_{\mathrm{render}}$ with the real image of every frame $t\in\mathcal{N}_{a}$ in which the instance is observed, where $a$ is the anchor frame of the target instance. Using the ego poses, the matched partial 3D points from each frame are transformed into the anchor frame coordinate system and aggregated as
\begin{equation}
\mathcal{C}^{\mathrm{agg}}_{a}
=
\bigcup_{t\in\mathcal{N}_{a}}
\left\{
\left(
\mathbf{p}^{\mathrm{amodal}}_{t,j},
E_{t\rightarrow a}\mathbf{p}^{\mathrm{partial}}_{t,j}
\right)
\right\}_{j=1}^{L_t},
\end{equation}
where $E_{t\rightarrow a}$ is the relative ego-pose transformation from frame $t$ to the anchor frame $a$. The final alignment for static instances is then estimated once from the aggregated correspondence set $\mathcal{C}^{\mathrm{agg}}_{a}$.
For dynamic instances, independently estimating a per-frame yaw-constrained similarity transform, including scale, yaw, and translation, leads to frame-to-frame scale jitter, since partial visibility and sparse LiDAR support can change substantially across frames. We therefore first estimate the metric scale at the anchor frame $a$ together with its yaw and translation as 
\begin{equation}
(s^\star,\theta_a^\star,\mathbf{t}_a^\star)
=
\arg\min_{s>0,\theta,\mathbf{t}}
\sum_{j=1}^{L_a}
\rho\!\left(
\left\|
\mathbf{p}^{\mathrm{partial}}_{a,j}
-
\left(
s R_{\mathrm{yaw}}(\theta)\mathbf{p}^{\mathrm{amodal}}_{a,j}
+
\mathbf{t}
\right)
\right\|_2
\right)
+
\lambda (s-s_0)^2,
\label{eq:eq9}
\end{equation}
where $s_0$ is the category-level scale prior obtained following OpenBox~\cite{openbox}, and $\rho(\cdot)$ is the robust penalty induced by the RANSAC inlier criterion. We then keep $s^\star$ fixed for the remaining frames and update only the yaw and translation at frame $\tau$ as 
\begin{equation}
(\theta_\tau,\mathbf{t}_\tau)
=
\arg\min_{\theta,\mathbf{t}}
\sum_{j=1}^{L_\tau}
\rho\!\left(
\left\|
\mathbf{p}^{\mathrm{partial}}_{\tau,j}
-
\left(
s^\star R_{\mathrm{yaw}}(\theta)\mathbf{p}^{\mathrm{amodal}}_{\tau,j}
+
\mathbf{t}
\right)
\right\|_2
\right).
\label{eq:eq10}
\end{equation}
The resulting alignment is written as
\begin{equation}
\mathbf{p}^{\mathrm{partial}}_{\tau,j}
\approx
s^\star R_{\mathrm{yaw}}(\theta_\tau)\mathbf{p}^{\mathrm{amodal}}_{\tau,j}
+
\mathbf{t}_\tau.
\end{equation}
We then apply the closeness-to-edge algorithm~\cite{lshape} to tightly fit a bounding box around the asset aligned in world space.

\section{Experiments}
\label{sec:experiments}

\subsection{Experimental Setup}
\noindent\textbf{Dataset \& Implementation Details}
We evaluate on two real-world driving datasets, nuScenes~\cite{nuscenes} with 700/150 train/val scenes and 10 classes, and WOD~\cite{waymo} with 798/202 train/val scenes and 3 classes. We train and evaluate over the full 360-degree field of view on nuScenes~\cite{nuscenes}, and only within the camera frustum on WOD~\cite{waymo}, following~\cite{openbox}.
Annotating the training split takes about 50 h on nuScenes and 156 h on WOD with 4 NVIDIA B200 GPUs (Appendix~\ref{appendix:runtime}).
For the 3D object detector, we use TransFusion~\cite{transfusion} and CenterPoint~\cite{centerpoint} on nuScenes~\cite{nuscenes}, and Voxel R-CNN~\cite{voxelrcnn} on WOD~\cite{waymo}. TransFusion and Voxel R-CNN are implemented in OpenPCDet~\cite{openpcdet}, and CenterPoint in MMDetection3D~\cite{mmdet}. We refer readers to Appendix~\ref{sec:appendix:implementation} for additional training details.\\
\vspace{-20pt}
\subsection{Main Result}
Following~\cite{upvl,openbox}, we assess \Ours{} under two scenarios. In Scenario 1, we train 3D object detectors with automated annotations and test them on the human-annotated validation set (\Cref{tab:table1,tab:table2-1,tab:table3}),
and in Scenario 2, we directly compare the annotations with human labels on the training
set (\Cref{tab:table2-2,tab:table4,tab:table5}).

\noindent\textbf{Results on nuScenes}
As shown in~\Cref{tab:table1}, \Ours{} substantially improves both mAP and NDS over prior methods, without additional strategies such as data augmentation~\cite{annofreeod,findnpropagate,opensight} or detector architecture modification~\cite{ovscan,upvl}, and improves AP for 9 out of 10 classes.
We attribute this broad improvement to the amodal shape priors used for box generation.
Under partial visibility, object points are often sparse or fragmented, so clustering-based post-processing~\cite{dbscan,hdbscan} either misses objects or fits boxes only to their visible regions, whereas \Ours{} generates boxes from the recovered amodal extent even when the visible points cover only a small portion of the object.
The large gains on bus, construction vehicle, and trailer support this explanation, as these large objects are the most affected by visible-only box fitting under partial observation.
The only exception is bicycle, which is known to be challenging in nuScenes~\cite{nuscenes} due to its thin structure and few LiDAR returns, making the partial observations less reliable for amodal alignment.

\begin{table}[!t]
\centering
\setlength{\tabcolsep}{3.5pt}
\caption{
    \textbf{3D object-detection results on the nuScenes~\cite{nuscenes} validation set.}
    Ped., M.C., Barr., C.V., and Trail. denote pedestrian, motorcycle, barrier, construction vehicle, and trailer, respectively.
    $^{\dagger}$ is trained on~\cite{transfusion} without semantic alignment head in OV-SCAN~\cite{ovscan}. 
    \textbf{Bold} and \underline{underline} indicate the best and the second-best results.}
\label{tab:table1}
\footnotesize
\vspace{+3pt}
\begin{tabular}{lcccccccccccc}
\toprule
Method & mAP$\uparrow$ & NDS$\uparrow$ & Car & Ped. & Truck & M.C. & Bicycle & Cone & Bus & Barr. & C.V. & Trail. \\
\midrule
UP-VL~\cite{upvl}          & 11.3 & 17.2 & 19.4 & 30.8 & \padnum{7.9} & 16.8 & 14.1 & 15.6 & \padnum{3.1} & \padnum{0.5} & \padnum{4.3} & \padnum{0.4} \\
Find n' Prop.~\cite{findnpropagate}   & 16.7 & 22.4 & 24.3 & 22.8 & \padnum{8.6} & 35.8 & \textbf{34.7} & 21.3 & 11.1 & \padnum{4.4} & \padnum{4.1} & \padnum{0.1} \\
OpenSight~\cite{opensight}       & 22.9 & 23.3 & 25.3 & 52.5 & 11.6 & 30.4 & 26.1 & 25.6 & \padnum{5.1} & \underline{42.2} & \padnum{\underline{8.7}} & \padnum{\underline{0.8}} \\
AnnofreeOD~\cite{annofreeod} & 22.4 & 26.4 & 50.2 & 51.2 & 17.1 & 27.1 & 13.4 & 17.1 & \underline{26.1} & \padnum{9.6} & \padnum{2.2} & \padnum{0.0} \\ 
OV-SCAN~\cite{ovscan}             & \underline{31.1} & \underline{32.8} & 61.6 & \underline{60.1} & \underline{30.3} & \underline{39.8} & \underline{31.0} & \underline{39.6} & 22.0 & 18.8 & \padnum{6.8} & \padnum{0.6} \\
OV-SCAN$^{\dagger}$~\cite{ovscan} & 28.7 & 31.4 & \underline{62.2} & 59.1 & 29.8 & 29.5 & 26.3 & 34.6 & 19.9 & 17.3 & \padnum{7.1} & \padnum{0.7} \\
\textbf{AmodalBox (Ours)} & \textbf{42.8} & \textbf{37.6} & \textbf{68.4} & \textbf{63.2} & \textbf{38.5} & \textbf{44.9} & 29.4 & \textbf{49.3} & \textbf{51.4} & \textbf{60.4} & \textbf{15.3} & \padnum{\textbf{7.2}} \\
\bottomrule
\end{tabular}
\vspace{-3pt}
\end{table}

\begin{table}[!t]
\centering
\vspace{-5pt}
\caption{
\textbf{3D object-detection results on WOD~\cite{waymo}.} All values denote $\mathrm{AP}_{\mathrm{3D}}$. $^{*}$ denotes models trained on the union of OpenBox~\cite{openbox} and \Ours{} annotations after removing duplicates with class-wise NMS. \textbf{Bold} and \underline{underline} indicate the best and second-best results.
}
\label{tab:table2}
\renewcommand{\arraystretch}{1.0}
\small
\setlength{\tabcolsep}{3pt}

\begin{subtable}[t]{0.49\textwidth}
\centering
\caption{Scenario 1}
\label{tab:table2-1}
\resizebox{\textwidth}{!}{%
\begin{tabular}{@{}l|ccc@{}}
\toprule
Method 
& \makecell{Vehicle \\ $\mathrm{IoU}_{0.5}$ / $\mathrm{IoU}_{0.7}$}
& \makecell{Pedestrian \\ $\mathrm{IoU}_{0.3}$ / $\mathrm{IoU}_{0.5}$}
& \makecell{Cyclist \\ $\mathrm{IoU}_{0.3}$ / $\mathrm{IoU}_{0.5}$} \\
\midrule
CPD~\cite{cpd}
& 30.30 / 20.90
& 14.28 / 11.22
& \padnum{3.47} / \padnum{3.08} \\

OpenBox~\cite{openbox}
& 70.49 / 32.41
& \underline{57.95} / \underline{17.11}
& \textbf{20.81} / \padnum{2.15} \\

\textbf{\Ours{}}
& \underline{82.94} / \underline{36.16}
& 55.92 / \padnum{6.41}
& 12.11 / \padnum{\underline{9.33}} \\

\textbf{\Ours{}$^{*}$}
& \textbf{84.37} / \textbf{47.88}
& \textbf{65.18} / \textbf{20.66}
& \underline{13.16} / \padnum{\textbf{9.63}} \\
\bottomrule
\end{tabular}
}
\end{subtable}
\hfill
\begin{subtable}[t]{0.49\textwidth}
\centering
\caption{Scenario 2}
\label{tab:table2-2}
\resizebox{\textwidth}{!}{%
\begin{tabular}{@{}l|ccc@{}}
\toprule
Method 
& \makecell{Vehicle \\ $\mathrm{IoU}_{0.5}$ / $\mathrm{IoU}_{0.7}$}
& \makecell{Pedestrian \\ $\mathrm{IoU}_{0.3}$ / $\mathrm{IoU}_{0.5}$}
& \makecell{Cyclist \\ $\mathrm{IoU}_{0.3}$ / $\mathrm{IoU}_{0.5}$} \\
\midrule
OpenBox~\cite{openbox}
& 26.38 / \padnum{5.68}
& \underline{38.53} / \textbf{\padnum{8.49}}
& \padnum{8.75} / \padnum{0.24} \\

\textbf{\Ours{}}
& \textbf{42.19} / \textbf{\padnum{9.21}}
& 26.96 / \padnum{1.66}
& \underline{\padnum{9.10}} / \textbf{\padnum{2.97}} \\

\textbf{\Ours{}$^{*}$}
& \underline{32.73} / \underline{\padnum{7.76}}
& \textbf{40.74} / \underline{\padnum{8.43}}
& \textbf{10.17} / \underline{\padnum{2.79}} \\

\bottomrule
\end{tabular}
}
\end{subtable}
\vspace{-15pt}
\end{table}

\begin{wraptable}{r}{0.46\textwidth}
\vspace{-12pt}
\centering
\small
\setlength{\tabcolsep}{4pt}
\caption{\textbf{Comparison with OpenBox~\cite{openbox} on nuScenes~\cite{nuscenes} under its 3-class protocol.} OpenBox numbers are as reported.}
\label{tab:table3}
\vspace{2pt}
\begin{tabular}{lccc}
\toprule
Method & Vehicle & Ped. & Cyclist \\
\midrule
OpenBox~\cite{openbox} & 40.9 & \textbf{62.7} & \padnum{5.2} \\
\Ours{} & \underline{59.5} & 46.7 & \underline{25.1} \\
\Ours{}$^{\ddagger}$ & \textbf{60.1} & \underline{62.0} & \textbf{26.6} \\
\bottomrule
\end{tabular}
\vspace{-8pt}
\end{wraptable}

\noindent\textbf{Comparison with OpenBox on nuScenes}
\Ours{} outperforms OpenBox~\cite{openbox} by a large margin on vehicles (+18.6 AP) and cyclists (+19.9 AP) but is lower on pedestrians.
As analyzed in Appendix~\ref{appendix:small-objects}, this gap stems almost entirely from extremely sparse instances. Among ground-truth pedestrians, 79\% have at most 15 LiDAR points, where too few correspondences remain for alignment, whereas with at least 16 points our recall reaches 96--99\%, and our overall precision is higher than that of OpenBox (94.9 vs.\ 78.8).
A density-conditioned fallback that tightly fits the observed points~\cite{openbox} only for such sparse pedestrians lifts pedestrian AP from 46.7 to 62.0 with marginal changes to the other classes (\Ours{}$^{\ddagger}$).

\noindent\textbf{Results on WOD} 
As shown in~\Cref{tab:table2}, \Ours{} improves the Vehicle class by a large margin at both IoU thresholds and the Cyclist class at IoU 0.5, while it falls behind OpenBox~\cite{openbox} for Pedestrian and for Cyclist at IoU 0.3. Vehicles benefit most because their full length is often unobserved, whereas OpenBox extends an initial box and selects the orientation of static objects by IoU, which becomes inaccurate under occlusion. Pedestrians are short and deformable, so they gain little from amodal shape recovery, and tightly fitting the observed points as in OpenBox can be more advantageous. The gain from mixing both annotations (\Ours{}$^{*}$) and the effect of the density-conditioned fallback on nuScenes (\Cref{tab:table3}) both support this interpretation.

\subsection{Ablation and Robustness Analysis}
We conduct the following analyses on 50 randomly sampled scenes from the nuScenes ~\cite{nuscenes} training set, whose sparse LiDAR observations provide a challenging setting for annotation, and evaluate the generated annotations directly against human labels.

\begin{table}[!t]
\caption{\textbf{Ablation studies on 50 nuScenes~\cite{nuscenes} training scenes.}
(a) Incremental ablation of the proposed modules.
(b) Correspondence source and category size prior; ``Vol.\ ratio'' is the box volume over the class-prior volume (p99: 99th percentile; $>\!20\times$: \% of boxes above $20\times$). Defaults in \textbf{bold}.}
\label{tab:table4}
\centering
\small
\begin{subtable}[t]{0.50\textwidth}
\centering
\caption{Module ablation}
\label{tab:table4-a}
\setlength{\tabcolsep}{3.5pt}
\resizebox{\linewidth}{!}{%
\begin{tabular}{lcccc}
\toprule
 & \makecell{Partial\\LiDAR} & \makecell{3D Asset Gen\\\& Matching} & \makecell{Motion-\\aware} & mAP$\uparrow$ / NDS$\uparrow$ \\
\midrule
(1) & \checkmark &            &            & \padnum{3.57} / \padnum{5.76} \\
(2) & \checkmark & \checkmark &            & 24.62 / 26.30 \\
(3) & \checkmark & \checkmark & \checkmark & \textbf{31.40} / \textbf{31.13} \\
\bottomrule
\end{tabular}}
\end{subtable}%
\hfill
\begin{subtable}[t]{0.47\textwidth}
\centering
\caption{Correspondence and scale prior}
\label{tab:table4-b}
\setlength{\tabcolsep}{3pt}
\resizebox{\linewidth}{!}{%
\begin{tabular}{lcccrr}
\toprule
 & & & & \multicolumn{2}{c}{Vol.\ ratio} \\
\cmidrule(lr){5-6}
Correspondence & $\lambda$ & mAP$\uparrow$ & NDS$\uparrow$ & p99$\downarrow$ & $>\!20\times$ \\
\midrule
ICP~\cite{icp}            & 1     & 21.80 & 20.72 & 15.88 & 0.54 \\
\textbf{Dense matching}   & \textbf{1} & \textbf{31.40} & \textbf{31.13} & \padnum{4.41} & 0.05 \\
Dense matching            & 0     & 31.99 & 31.65 & 12.65 & 0.72 \\
Dense matching            & 0.1   & 32.24 & 31.93 & \padnum{7.57} & 0.23 \\
Dense matching            & 10    & 29.77 & 29.03 & \padnum{3.30} & 0.00 \\
\bottomrule
\end{tabular}}
\end{subtable}
\vspace{-12pt}
\end{table}

\begin{table}[!t]
\caption{\textbf{Robustness analysis on 50 nuScenes~\cite{nuscenes} training scenes.}
(a) Each corruption is applied to the generated assets. Cov.\ is recall at a 2\,m center distance. (b) Oracle states come from human annotation tracks. Flips are applied to the oracle states. D$\rightarrow$S flips the given fraction of dynamic tracks to static (S$\rightarrow$D the reverse).}
\label{tab:table5}
\centering
\small
\begin{subtable}[t]{0.50\textwidth}
\centering
\caption{3D asset quality sensitivity}
\label{tab:table-a}
\setlength{\tabcolsep}{3pt}
\resizebox{\linewidth}{!}{%
\begin{tabular}{lcccc}
\toprule
Asset corruption & mAP$\uparrow$ & Cov.$\uparrow$ & mATE$\downarrow$ & mASE$\downarrow$ \\
\midrule
Amodal half removed & 19.3 & 41.2 & 0.826 & 0.708 \\
Same-class swap & 21.2 & 39.7 & 0.658 & 0.609 \\
Appearance flattening & 24.8 & 43.5 & 0.593 & 0.538 \\
Anisotropic extent (30\%) & 29.0 & 50.7 & 0.597 & 0.597 \\
\textbf{None (ours)} & \textbf{31.4} & \textbf{51.1} & \textbf{0.518} & \textbf{0.499} \\
\bottomrule
\end{tabular}}
\end{subtable}%
\hfill
\begin{subtable}[t]{0.48\textwidth}
\centering
\caption{Motion-state sensitivity}
\label{tab:table5-b}
\setlength{\tabcolsep}{2.5pt}
\resizebox{\linewidth}{!}{%
\begin{tabular}{lcc}
\toprule
Motion state & mAP$\uparrow$ & NDS$\uparrow$ \\
\midrule
Estimated (ours) & 31.40 & 31.13 \\
Oracle & \textbf{33.50} & \textbf{32.01} \\
D$\rightarrow$S 10/20/40\% & 32.79/31.49/30.04 & 31.78/31.09/30.55 \\
S$\rightarrow$D 10/20/40\% & 33.08/33.16/32.51 & 31.72/31.91/31.27 \\
\bottomrule
\end{tabular}}
\end{subtable}
\vspace{-20pt}
\end{table}

\noindent\textbf{Module Ablation} We analyze the effect of each proposed module by adding each module incrementally, as shown in~\Cref{tab:table4-a}. Comparing (1) and (2), mAP and NDS increase by 21.1 and 20.5 points, respectively. This indicates that relying only on partial LiDAR point clouds tends to produce mis-sized boxes, whereas using amodal priors enables more reliable box annotation, improving both mAP and NDS, which reflects localization and size quality. 
Comparing (2) and (3), we also observe additional performance gains. 
We conjecture that repeatedly generating boxes for the same instance in each frame can cause box-size jittering, especially for dynamic objects, because the generated amodal priors may vary across frames. 
For static instances, aggregating points across frames makes the alignment more robust to outliers and improves box accuracy.

\noindent\textbf{Correspondence and Scale Prior}
\Cref{tab:table4-b} examines the two design choices inside 3D asset generation and matching.
Replacing dense matching with geometry-only ICP~\cite{icp}, initialized with the same yaw candidates and size prior, reduces mAP by 9.6 and inflates the tail of the box-volume distribution (p99 from 4.41 to 15.88), which is the local-minimum failure that dense matching avoids.
The size prior in Eq.~\cref{eq:eq9} does not drive the accuracy gain, since removing it ($\lambda=0$) gives 32.0 mAP while over-weighting it ($\lambda=10$) hurts, but it suppresses rare scale divergence that the center-distance-based nuScenes AP is blind to, reducing the fraction of boxes above $20\times$ the class volume from 0.72\% to 0.05\%, so we keep $\lambda=1$ as a safeguard.

\noindent\textbf{Robustness}
\Cref{tab:table5} examines how errors in the generated assets and in the estimated motion states propagate through the pipeline.
Appearance corruptions, such as flattened textures or an asset swapped with another instance of the same class, mainly reduce coverage, which indicates that dense matching rejects assets inconsistent with the image rather than propagating them into boxes.
Geometric corruptions pass through this gate because matching verifies appearance rather than metric extent, so a 30\% anisotropic distortion keeps coverage but raises mASE, and removing half of the amodal geometry is the most harmful case since it discards the unobserved extent that the method relies on.
Motion-state errors show a similar asymmetry, as misclassifying dynamic tracks as static costs 3.5 mAP at a 40\% error rate whereas the reverse costs only 1.0, because the static branch aggregates the correspondences of a moving object across frames.

\begin{figure}[!t]
    \centering
    \includegraphics[width=0.85\linewidth]{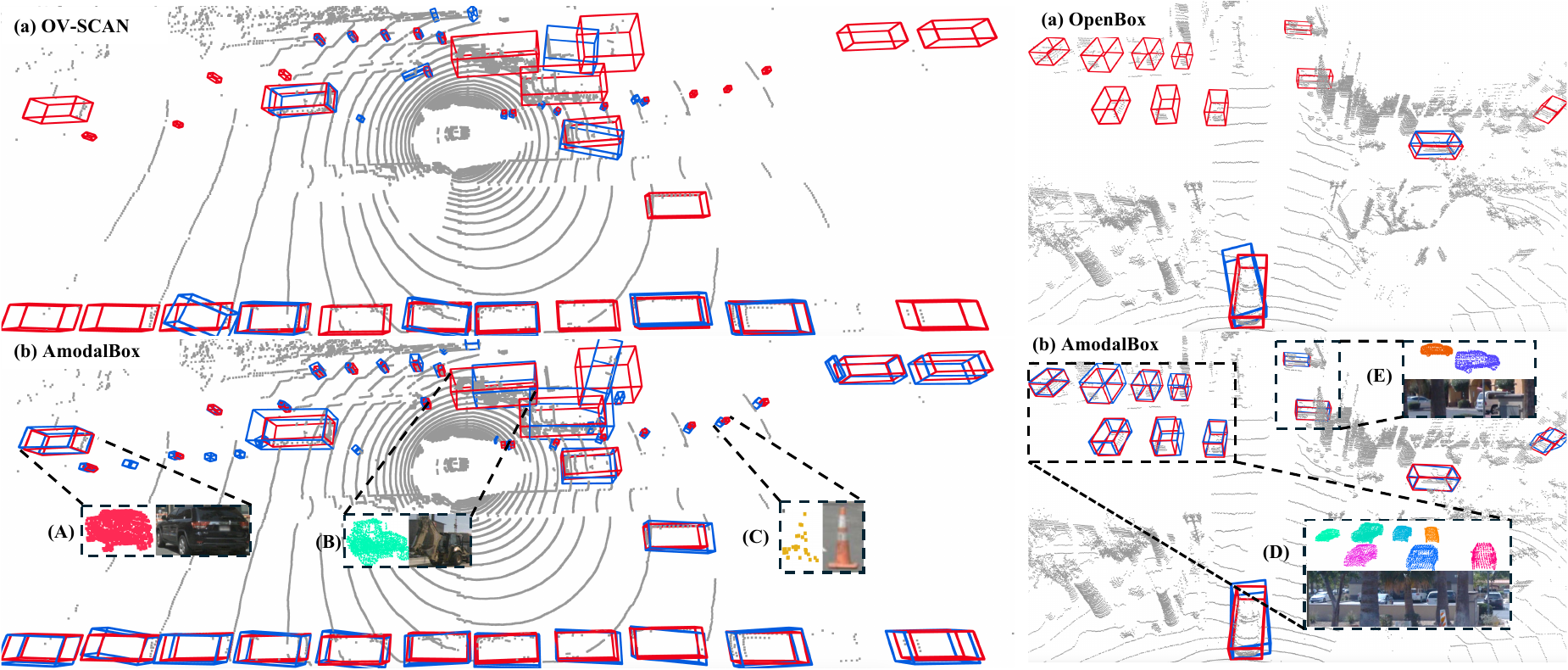}
    \caption{\textbf{Visualization of annotations} on nuScenes~\cite{nuscenes} (left) and WOD~\cite{waymo} (right). Red boxes are human annotations, blue boxes are automated annotations, and the colored (non-gray) points are the aligned amodal asset. Best viewed in color and zoomed in.}
    \label{fig:figure4}
\end{figure}

\subsection{Qualitative Result}
\Cref{fig:figure4} compares \Ours{} with OV-SCAN~\cite{ovscan} on nuScenes~\cite{nuscenes} and with OpenBox~\cite{openbox} on WOD~\cite{waymo}.
On nuScenes~\cite{nuscenes}, \Ours{} achieves higher recall and more accurate box localization than OV-SCAN~\cite{ovscan}. In (A) and (B), OV-SCAN~\cite{ovscan} fails to produce boxes due to its selective alignment strategy, whereas \Ours{} annotates these objects by recovering their amodal shapes. In (C), \Ours{} correctly annotates the static traffic cone by explicitly distinguishing the instance motion state.
On WOD~\cite{waymo}, in (D) and (E), OpenBox~\cite{openbox} produces no annotations despite the presence of points, because the points are removed during its context-aware refinement, whereas \Ours{} first aligns an amodal asset to the observed point cloud and annotates these objects.

\section{Conclusion}
\label{sec:conclusion}
We proposed \Ours{}, an automated open-vocabulary 3D bounding box annotation pipeline that addresses partial visibility in driving scenes by aligning partial world-space LiDAR observations with amodal 3D assets generated from images in a motion-aware manner.
Experiments on two datasets demonstrate that \Ours{} operates more effectively than prior methods under partial visibility, and our ablation and robustness analyses examine the effect of each module and the propagation of errors across stages.
We believe that \Ours{} will advance research toward more robust 3D annotation pipelines.  
\\\noindent\textbf{Limitations}
Our method has several limitations. Since \Ours{} relies on VFMs outputs, reflected objects on glass surfaces introduce false positives. In addition, adverse weather can degrade 2D matching and lead to incorrect 3D alignment, while inaccurate scene flow may misclassify dynamic objects as static, causing false positives to persist across frames. Finally, small, thin, or deformable objects with few LiDAR returns benefit little from amodal alignment.

\acknowledgments{This work was supported by IITP grant (RS-2026-25517417: Development of Compression, Reconstruction, and Rendering Technologies for Free-viewpoint Media (30\%), RS-2025-02303703: Realworld Multi-space Fusion and 6DoF Free-viewpoint Immersive Visualization for Extended Reality (30\%), RS-2025-25442338: AI Star Fellowship Support Program (Seoul National University) (35\%), RS-2021-II211343: AI Graduate School Program (Seoul National University) (5\%)) funded by the Korea government (MSIT). 
}

\bibliography{main}  %
\clearpage
\appendix
\section{Appendix}
\subsection{Motion-aware Alignment}
\label{appendix:motion-aware}
\noindent\textbf{Problem Formulation} 
Given the 3D-3D correspondence set $\mathcal{C}_t=\{(\mathbf{p}^{\mathrm{amodal}}_{t,j},\mathbf{p}^{\mathrm{partial}}_{t,j})\}_{j=1}^{L_t}$ defined in Sec.~\ref{method/sub/3.2}, we estimate the transformation from the canonical amodal asset to the world-space partial observation. 
Since most object categories can be reasonably approximated with only a yaw rotation, we formulate the alignment as a yaw-constrained, scale-regularized 5-DoF problem:
\begin{equation}
\mathbf{p}^{\mathrm{partial}}_{t,j}
\approx
s\,R_{\mathrm{yaw}}(\theta)\,\mathbf{p}^{\mathrm{amodal}}_{t,j} + \mathbf{t},
\end{equation}
where $s>0$ is the scale, $R_{\mathrm{yaw}}(\theta)\in SO(3)$ is the yaw-only rotation matrix, and $\mathbf{t}\in\mathbb{R}^3$ is the translation. To estimate the yaw-only rotation, we project the 3D correspondences onto the BEV axes and solve a 2D Procrustes problem.

\noindent\textbf{Hypothesis Generation} 
Let $\bar{\mathbf{p}}^{\mathrm{amodal}}_j \in \mathbb{R}^2$ and $\bar{\mathbf{p}}^{\mathrm{partial}}_j \in \mathbb{R}^2$ denote the corresponding BEV-plane coordinates of $\mathbf{p}^{\mathrm{amodal}}_j$ and $\mathbf{p}^{\mathrm{partial}}_j$, respectively. We first compute their centroids
\begin{equation}
\mu_{\mathrm{amodal}} = \frac{1}{L_t}\sum_{j=1}^{L_t}\bar{\mathbf{p}}^{\mathrm{amodal}}_{t,j},
\qquad
\mu_{\mathrm{partial}} = \frac{1}{L_t}\sum_{j=1}^{L_t}\bar{\mathbf{p}}^{\mathrm{partial}}_{t,j},
\end{equation}
and the centered coordinates
\begin{equation}
\tilde{\mathbf{p}}^{\mathrm{amodal}}_{t,j}
=
\bar{\mathbf{p}}^{\mathrm{amodal}}_{t,j} - \mu_{\mathrm{amodal}},
\qquad
\tilde{\mathbf{p}}^{\mathrm{partial}}_{t,j}
=
\bar{\mathbf{p}}^{\mathrm{partial}}_{t,j} - \mu_{\mathrm{partial}}.
\end{equation}
The BEV cross-covariance matrix is then
\begin{equation}
C = \frac{1}{L_t}\sum_{j=1}^{L_t}
\tilde{\mathbf{p}}^{\mathrm{amodal}}_{t,j}
\tilde{\mathbf{p}}^{\mathrm{partial}\top}_{t,j}.
\end{equation}
Using the singular value decomposition $C = U\Sigma V^\top$, the BEV rotation is given by
\begin{equation}
R_{\mathrm{bev}} = V
\begin{bmatrix}
1 & 0\\
0 & \det(VU^\top)
\end{bmatrix}
U^\top.
\end{equation}

The BEV translation is computed as
\begin{equation}
\mathbf{t}_{\mathrm{bev}}
=
\mu_{\mathrm{partial}} - s\,R_{\mathrm{bev}}\,\mu_{\mathrm{amodal}},
\end{equation}
while the vertical translation is estimated robustly using the median
\begin{equation}
t_z
=
\operatorname{median}_j
\left(
\left[\mathbf{p}^{\mathrm{partial}}_{t,j}\right]_z - s\,\left[\mathbf{p}^{\mathrm{amodal}}_{t,j}\right]_z
\right).
\end{equation}

\noindent\textbf{Local Optimization} 
To regularize the scale, we further introduce a class-dependent prior. Let $(l,w,h)_{\mathrm{class}}$ denote the coarse class-size prior and let $d_{\mathrm{src}}$ be the diagonal length of the robust extent of the Gaussian centers in $A^k$. We define the prior scale as
\begin{equation}
d_{\mathrm{src}} = \left\| \operatorname{extent}_{q_{\mathrm{low}}, q_{\mathrm{high}}}(A^k) \right\|_2,
\qquad
d_{\mathrm{prior}} = \|(l,w,h)_{\mathrm{class}}\|_2,
\end{equation}
\begin{equation}
s_0 = \frac{d_{\mathrm{prior}}}{d_{\mathrm{src}}}.
\end{equation}

Instead of relying solely on the data-driven scale estimate, we combine it with the class prior:
\begin{equation}
s \approx \frac{s_{\mathrm{data}} + \lambda s_0}{1+\lambda},
\end{equation}
where $\lambda$ controls the prior strength. We solve this problem using LO-RANSAC~\cite{loransac} with closed-form alignment estimation inspired by Umeyama~\cite{umeyama}. Overall, this can be viewed as a robust approximation to
\begin{equation}
\min_{s,\theta,\mathbf{t}}
\sum_j
\rho\!\left(
\left\|
\mathbf{p}^{\mathrm{partial}}_{t,j}
-
\left(
s\,R_{\mathrm{yaw}}(\theta)\,\mathbf{p}^{\mathrm{amodal}}_{t,j} + \mathbf{t}
\right)
\right\|_2
\right)
+
\lambda (s-s_0)^2,
\end{equation}
where $\rho(\cdot)$ is induced implicitly by the RANSAC inlier criterion.

\newpage

\subsection{Rendered View Selection}
\label{appendix:view-selection}
We select the rendered view by constructing a render bank around the initial object orientation predicted by~\cite{fastsam3d}. Specifically, let $\theta_{\mathrm{initial}}$ denote the initial yaw angle estimated by ~\cite{fastsam3d}. For each candidate, we apply an additional yaw offset $\Delta\theta$ around the vertical axis and render the asset at yaw $\theta_{\mathrm{initial}}+\Delta\theta$. Each rendered candidate produces RGB-D images, which are matched to the real image using RoMa~\cite{roma}. The resulting 2D correspondences are lifted to 3D to form 3D-3D correspondences and are evaluated through robust alignment. We select the candidate render view that yields the largest number of RANSAC inliers. For static objects, we use a narrower offset set,
$\Delta\theta \in \{-30^\circ,-15^\circ,0^\circ,15^\circ,30^\circ\}$,
and aggregate correspondences from multiple frames in the anchor ego frame before the final alignment. For dynamic objects, we use a wider offset set,
$\Delta\theta \in \{-45^\circ,-22.5^\circ,0^\circ,22.5^\circ,45^\circ\}$,
to account for larger pose uncertainty. After selecting the anchor render view, subsequent frames are aligned while keeping the anchor scale fixed. Only nearby yaw-offset bins are searched to preserve temporal consistency.

\subsection{Category-level Size Priors}
\label{appendix:size-prior}
Following OpenBox~\cite{openbox}, we used ChatGPT-5.5~\cite{chatgpt2026} to estimate category-level 3D bounding box size priors. Specifically, for each object category, we queried the model with the following prompt and used the returned dimensions as approximate class-wise priors.

\newenvironment{promptbox}
  {\begin{quote}\ttfamily\small\raggedright\setlength{\parindent}{0pt}\setlength{\parskip}{2pt}}
  {\end{quote}}
\newenvironment{responsebox}
  {\begin{quote}\ttfamily\small\raggedright\setlength{\parindent}{0pt}\setlength{\parskip}{2pt}}
  {\end{quote}}

\newcommand{\Prompt}{\textsf{\textbf{Prompt}}}
\newcommand{\Response}{\textsf{\textbf{Response}}}

\begin{promptbox}
\Prompt: What are the typical 3D bounding box dimensions for \texttt{[class]}?
\end{promptbox}

\begin{responsebox}
\Response: The typical 3D bounding box size for \texttt{[class]} is approximately given as Length~$\times$~Width~$\times$~Height in meters, based on common autonomous driving datasets such as nuScenes, KITTI, and the Waymo Open Dataset.
\end{responsebox}

\begin{table}[!t]
\centering
\caption{Training configurations for 3D object detectors.}
\label{tab:table6}
\begin{tabular}{lcc}
\toprule
Config & Voxel R-CNN~\cite{voxelrcnn} & TransFusion~\cite{transfusion} \\
\midrule
Optimizer           & \texttt{AdamW}        & \texttt{AdamOneCycle} \\
Base learning rate  & $1\mathrm{e}{-2}$     & $1\mathrm{e}{-3}$ \\
Weight decay        & $1\mathrm{e}{-3}$     & $1\mathrm{e}{-2}$ \\
Momentum            & $0.9$                 & $0.9$ \\
Momentum range      & $[0.95, 0.85]$        & $[0.9, 0.8052631]$ \\
Learning rate decay & $0.1$                 & $0.1$ \\
Learning rate clip  & $1\mathrm{e}{-7}$     & $1\mathrm{e}{-7}$ \\
Gradient norm clip  & $10$                  & $35$ \\
Batch size          & $16$                  & $16$ \\
Epoch               & $20$                  & $20$ \\
\bottomrule
\end{tabular}
\end{table}
\begin{table}[t]
\centering
\caption{Hyperparameters for yaw-constrained scale-regularized LO-RANSAC alignment.}
\label{tab:alignment_hyperparams}
\begin{tabular}{lll}
\toprule
Hyperparameter & Value & Description \\
\midrule
$q_{\mathrm{low}}$ & 0.05 & Lower quantile for robust source extent \\
$q_{\mathrm{high}}$ & 0.95 & Upper quantile for robust source extent \\
$\lambda$ & 1.0 & Strength of the class-size scale prior \\
Inlier threshold & 0.3 m & RANSAC inlier distance threshold \\
Minimum inliers & 10 & Minimum inliers for accepting a hypothesis \\
Number of iterations & 1000 & Maximum LO-RANSAC iterations \\
Minimum iterations & 1000 & Minimum iterations before stopping \\
Hypotheses per step & 64 & Number of hypotheses evaluated per batch \\
LO iterations & 5 & Number of local optimization refinements \\
LO minimum inliers & 10 & Minimum inliers required for local optimization \\
\bottomrule
\end{tabular}
\end{table}
\subsection{Implementation details }
\noindent\textbf{Training and 3D Object Detector Details} 
For a fair comparison, we train Voxel R-CNN~\cite{voxelrcnn} on WOD~\cite{waymo}, following~\cite{cpd,openbox}, and TransFusion~\cite{transfusion} on nuScenes~\cite{nuscenes}, following~\cite{ovscan}. We use the default OpenPCDet~\cite{openpcdet} configurations shown in~\Cref{tab:table6}, without modifying any hyperparameters.\\
\noindent\textbf{Automated Annotation Details} 
\label{sec:appendix:implementation}
The hyperparameters used for motion-aware alignment are listed in~\Cref{tab:alignment_hyperparams}. In addition, for the mixed annotation with NMS used in ~\Cref{tab:table2-2}, we use the NMS implementation provided by OpenPCDet~\cite{openpcdet}. The NMS threshold is set to 0.01, and NMS is performed in a class-wise manner.

\subsection{Small, Deformable, and Thin Objects}
\label{appendix:small-objects}
\begin{table}[!t]
\centering
\small
\caption{\textbf{Analysis of small, deformable, and thin objects on nuScenes~\cite{nuscenes}.}
(a) Quality of the raw \Ours{} annotations per super-class (Cov.: recall of human boxes at 2\,m; Prec.: precision; ATE/ASE: translation and scale errors of matched boxes), next to the detector AP of~\Cref{tab:table3}.
(b) Pedestrian recall at 2\,m as a function of the number of LiDAR points inside the human box. ``GT'' is the share (\%) of ground-truth pedestrians in each bin.}
\label{tab:ped}
\begin{subtable}[t]{0.50\textwidth}
\centering
\caption{Per-class annotation quality}
\label{tab:ped-a}
\setlength{\tabcolsep}{4pt}
\resizebox{\linewidth}{!}{%
\begin{tabular}{lrrrrr}
\toprule
Class & Cov.$\uparrow$ & Prec.$\uparrow$ & ATE$\downarrow$ & ASE$\downarrow$ & Det.\ AP$\uparrow$ \\
\midrule
Vehicle    & 62.3 & 81.0 & 0.35 & \textbf{0.25} & 59.5 \\
Pedestrian & 47.8 & \textbf{94.9} & \textbf{0.26} & 0.40 & 46.7 \\
Cyclist    & 60.7 & 82.3 & 0.31 & 0.48 & 25.1 \\
\bottomrule
\end{tabular}}
\end{subtable}%
\hfill
\begin{subtable}[t]{0.46\textwidth}
\centering
\caption{Pedestrian recall by LiDAR density}
\label{tab:ped-b}
\setlength{\tabcolsep}{4pt}
\resizebox{\linewidth}{!}{%
\begin{tabular}{lrrr}
\toprule
LiDAR points & GT (\%) & \Ours{} & OpenBox~\cite{openbox} \\
\midrule
1--5     & 46 & 15.6 & \textbf{40.3} \\
6--15    & 33 & 63.0 & \textbf{66.2} \\
16--40   & 12 & \textbf{96.3} & 74.3 \\
$>$40    & \padnum{9} & \textbf{98.6} & 67.9 \\
\midrule
Precision & -- & \textbf{94.9} & 78.8 \\
\bottomrule
\end{tabular}}
\end{subtable}
\end{table}

\Cref{tab:ped-a} evaluates the raw annotations per super-class on nuScenes. The two weaker classes fail in different ways. Pedestrians are recall-limited, in that the boxes we do produce are the most accurate of all classes (highest precision, lowest ATE), but coverage is the lowest. \Cref{tab:ped-b} locates the missing recall in extremely sparse instances, which make up 79\% of the ground-truth pedestrians. With at most 15 LiDAR points, too few correspondences survive the lifting step for a reliable alignment, whereas with at least 16 points \Ours{} recovers 96--99\% of the pedestrians, well above OpenBox~\cite{openbox}. Cyclists are instead scale-limited. Their coverage matches vehicles, but ASE is the largest, and a detector trained on them inherits the error, since the thin, non-convex bicycle-plus-rider geometry is poorly captured by a single generated asset.

\noindent\textbf{Density-conditioned fallback}
Motivated by this analysis, \Ours{}$^{\ddagger}$ in~\Cref{tab:table3} adds a single rule. For pedestrian instances with at most 15 LiDAR points, amodal alignment is skipped and the box is fitted tightly to the observed points with the category size prior, following~\cite{openbox}. This lifts pedestrian AP from 46.7 to 62.0 while leaving vehicle and cyclist AP essentially unchanged (59.5 to 60.1 and 25.1 to 26.6), so that \Ours{}$^{\ddagger}$ matches or exceeds OpenBox on all three classes.

\subsection{Runtime Breakdown}
\label{appendix:runtime}
\begin{table}[!t]
\centering
\small
\caption{\textbf{Annotation runtime breakdown} for the training splits on 4 NVIDIA B200 GPUs. ``Unit'' denotes the granularity at which each stage runs; asset generation is executed once per track rather than per frame.}
\label{tab:runtime}
\setlength{\tabcolsep}{8pt}
\begin{tabular}{llcc}
\toprule
Stage & Unit & nuScenes~\cite{nuscenes} & WOD~\cite{waymo} \\
\midrule
2D instance \& motion cues (\Cref{method/sub/3.1}) & frame & $\sim$4\,h & $\sim$18\,h \\
Asset generation \& rendering & track & $\sim$11\,h & $\sim$38\,h \\
Matching, alignment \& box fitting & track-frame & $\sim$35\,h & $\sim$100\,h \\
\midrule
Total & split & $\sim$50\,h & $\sim$156\,h \\
\bottomrule
\end{tabular}
\end{table}

\Cref{tab:runtime} breaks down the annotation time of the training splits by stage.
The 2D instance and motion cues are computed once per frame, and 3D asset generation and rendering are executed once per track at its anchor frame, so their cost is amortized over all frames of the track.
The remaining matching, alignment, and box fitting are performed for every frame of every track and account for roughly two thirds of the total time.
This stage is independent across instances and can be parallelized directly.

\subsection{Cross-view Instance Association}
\label{appendix:cross-view}
\begin{table}[!t]
\centering
\small
\caption{\textbf{Effect of cross-view association and duplicate removal on 50 nuScenes~\cite{nuscenes} training scenes.} ``Box/frame'' is the average number of annotated boxes per keyframe.}
\label{tab:dup}
\setlength{\tabcolsep}{8pt}
\begin{tabular}{lccc}
\toprule
Cross-view association \& NMS & mAP$\uparrow$ & NDS$\uparrow$ & Box/frame \\
\midrule
Off & 29.2 & 29.6 & 18.7 \\
\textbf{On (\Ours{})} & \textbf{31.4} & \textbf{31.1} & 13.8 \\
\bottomrule
\end{tabular}
\end{table}

Instance IDs produced by the tracker are consistent only within a camera view, whereas an object may be visible in two adjacent cameras simultaneously or move from one camera to another over time.
We therefore first remap the per-camera track IDs to be unique within a scene.
At each keyframe, tracks from different cameras are associated whenever their masks share a sufficient number of LiDAR points, and the associations are propagated over time, which covers both simultaneous multi-view observations and camera hand-offs.
Residual duplicate boxes are removed by class-wise BEV NMS that keeps the box enclosing more LiDAR points.
\Cref{tab:dup} shows that duplicate removal reduces the number of boxes by about a quarter (18.7 to 13.8 per keyframe) and improves mAP by 2.2 points.

\subsection{Failure Cases}
\begin{figure}[!t]
    \centering
    \includegraphics[width=\linewidth]{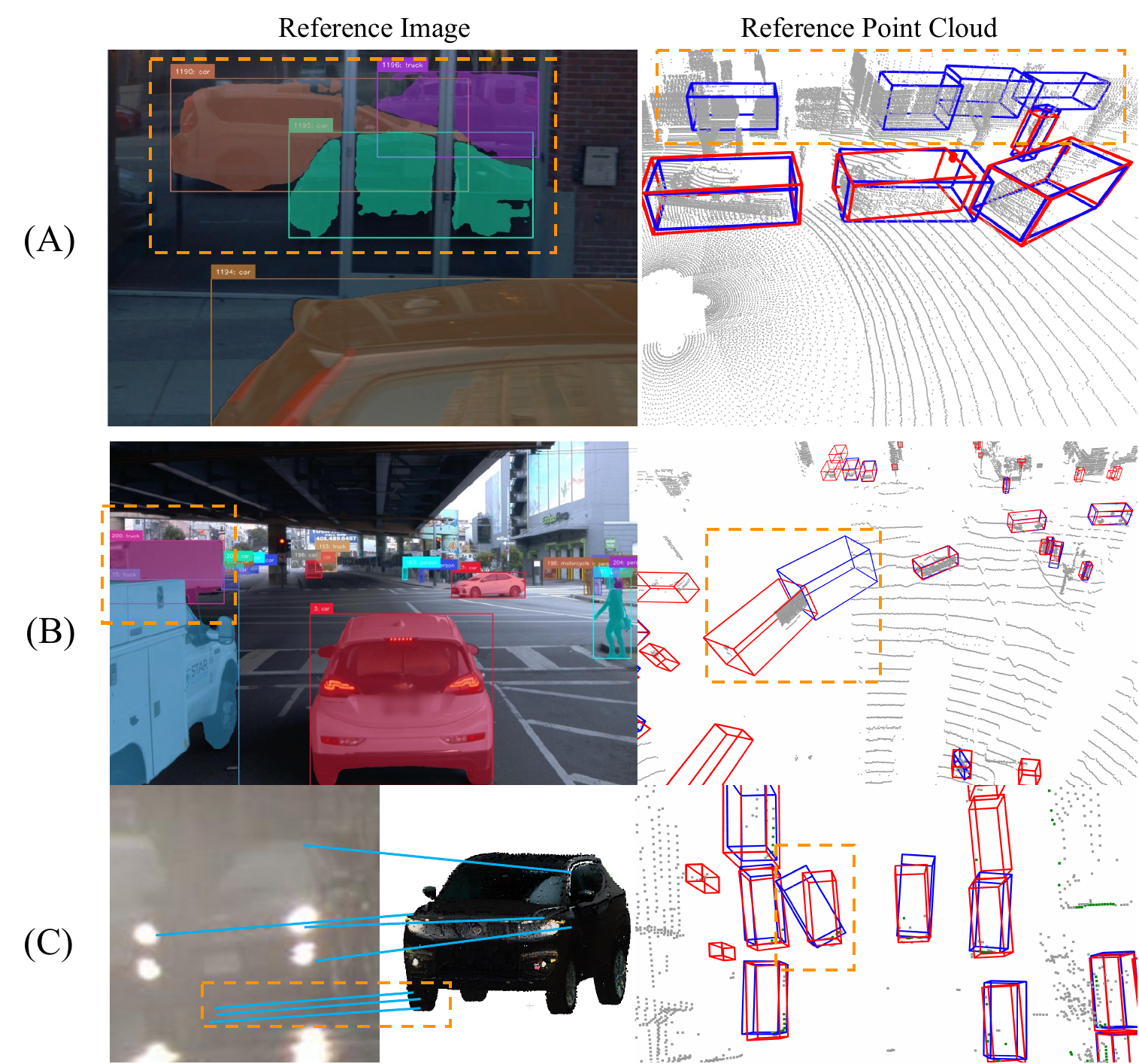}
    \caption{\textbf{Visualization of Failure Cases}. 
    Each row shows a representative failure case. 
    (A) illustrates a failure caused by vision foundation models (VFMs), while (B) shows a failure caused by incorrect object-motion estimation. 
    (C) shows a failure case in adverse weather, where inaccurate matching leads to erroneous 3D-3D alignment. 
    Red boxes denote human annotations, and blue boxes denote \Ours{}. 
    Best viewed in color and zoomed in.}
    \vspace{-10pt}
    \label{fig:figure5}
\end{figure}
As discussed in~\Cref{sec:conclusion}, the limitations of \Ours{} can be broadly categorized into three cases. In~\Cref{fig:figure5}, (A) shows a failure case caused by Grounding DINO~\cite{groundingdino}, where a vehicle reflected on glass is incorrectly detected. Since the point clouds inside the falsely detected box correspond to a flat glass surface, which resembles the side surface of a vehicle, the amodal 3D asset can still be aligned with the LiDAR point cloud, resulting in false-positive boxes. One possible solution is to adopt context-aware refinement, as in OpenBox~\cite{openbox}, to filter instance-level point clouds from the entire cluster. Another possible direction is to prompt background classes, as in UP-VL~\cite{upvl}, and suppress alignment when they overlap with foreground instances.

In (B), the failure is caused by inaccurate scene flow or PP score~\cite{modest} estimation, which leads to an incorrect prediction of the instance motion state. As shown by the orange box, a dynamic truck is temporarily misclassified as static, resulting in a false-positive box. To mitigate this issue, one could improve motion estimation by incorporating optical flow~\cite{raft} at the image level.

(C) shows a failure case under adverse weather conditions, where 2D-2D matching incorrectly associates the rear wheel in the real image with the front wheel in the rendered image, as highlighted by the orange box. This erroneous correspondence propagates to incorrect 3D-3D alignment. A possible remedy is to estimate correspondence reliability using local feature similarity, geometric consistency, and reprojection error, and discard low confidence matches before lifting them to 3D.

\begin{figure}[!t]
    \centering
    \includegraphics[width=0.8\linewidth]{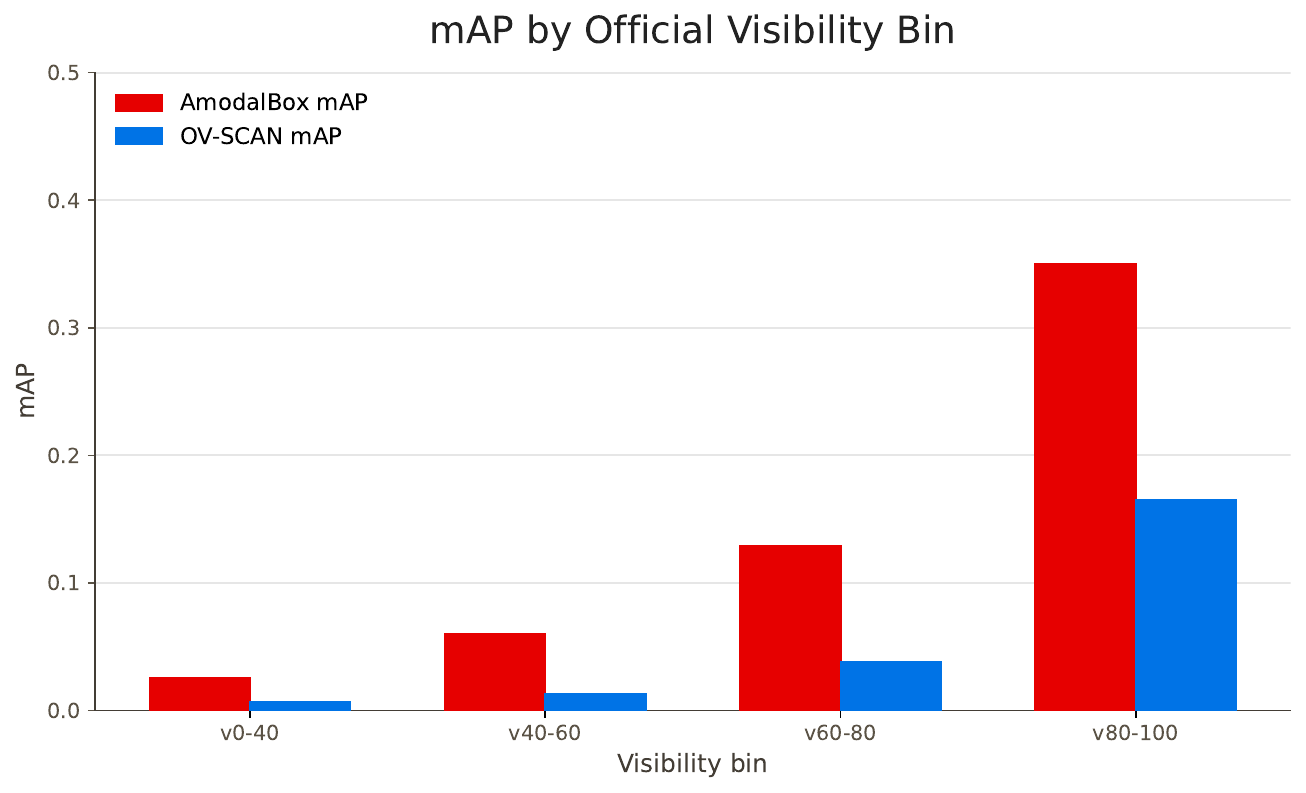}
     \caption{\textbf{Camera visibility-conditioned mAP on nuScenes~\cite{nuscenes}.} Camera visibility is defined using the official nuScenes~\cite{nuscenes} visibility annotations and grouped into four bins corresponding to 0-40\%, 40-60\%, 60-80\%, and 80-100\% visible object fraction.
    }
    \label{fig:figure6}
\end{figure}
\vspace{15pt}
\begin{table*}[!t]
\centering
\caption{
\textbf{Comparison by distance and LiDAR visibility on the nuScenes~\cite{nuscenes} validation set.}
We use the number of LiDAR points inside each human-annotated 3D box as a proxy for LiDAR visibility.
For each class, boxes with zero LiDAR points are excluded, and the remaining boxes are split into class-specific tertiles:
$(0, p_{33}]$ for \emph{Low}, $(p_{33}, p_{66}]$ for \emph{Mid}, and $(p_{66}, \infty)$ for \emph{High}.
Metrics are computed on each distance-visibility subset using the nuScenes detection protocol.
}
\label{tab:visibility}
\resizebox{\textwidth}{!}{
\begin{tabular}{lcccccccccccc}
\toprule
\multirow{3}{*}{Method}
& \multicolumn{12}{c}{mAP$\uparrow$} \\
\cmidrule(lr){2-13}
& \multicolumn{3}{c}{0-10m}
& \multicolumn{3}{c}{10-20m}
& \multicolumn{3}{c}{20-30m}
& \multicolumn{3}{c}{30-40m} \\
\cmidrule(lr){2-4}
\cmidrule(lr){5-7}
\cmidrule(lr){8-10}
\cmidrule(lr){11-13}
& Low & Mid & High
& Low & Mid & High
& Low & Mid & High
& Low & Mid & High \\
\midrule
OV-SCAN~\cite{ovscan}
& 2.18 & 4.75 & 54.15
& 3.05 & 11.67 & 43.04
& 2.88 & 18.05 & 27.46
& 2.63 & 15.41 & 10.45 \\
\Ours{}
& \textbf{7.37} & \textbf{11.27} & \textbf{64.92}
& \textbf{7.27} & \textbf{22.77} & \textbf{60.48}
& \textbf{6.55} & \textbf{32.20} & \textbf{46.77}
& \textbf{4.17} & \textbf{25.87} & \textbf{23.81} \\
\midrule
\multirow{3}{*}{Method}
& \multicolumn{12}{c}{NDS$\uparrow$} \\
\cmidrule(lr){2-13}
& \multicolumn{3}{c}{0-10m}
& \multicolumn{3}{c}{10-20m}
& \multicolumn{3}{c}{20-30m}
& \multicolumn{3}{c}{30-40m} \\
\cmidrule(lr){2-4}
\cmidrule(lr){5-7}
\cmidrule(lr){8-10}
\cmidrule(lr){11-13}
& Low & Mid & High
& Low & Mid & High
& Low & Mid & High
& Low & Mid & High \\
\midrule
OV-SCAN~\cite{ovscan}
& 10.48 & 15.24 & 47.34
& 13.81 & 22.67 & 37.61
& 12.96 & 23.33 & 28.21
& 13.36 & 19.50 & 16.04 \\
\Ours{}
& \textbf{18.51} & \textbf{17.64} & \textbf{51.44}
& \textbf{17.28} & \textbf{26.27} & \textbf{45.51}
& \textbf{16.51} & \textbf{30.90} & \textbf{38.86}
& \textbf{13.76} & \textbf{26.01} & \textbf{24.81} \\
\bottomrule
\end{tabular}
}
\end{table*}

\subsection{Analysis for partial visibility}
To further analyze the effectiveness of our method under partial visibility, we conduct modality-specific evaluations, as shown in~\Cref{tab:visibility} and ~\Cref{fig:figure6}. Since partial visibility is not explicitly defined for LiDAR, we first divide objects by distance. In addition, because annotation difficulty varies across classes depending on the number of observed LiDAR points, we further split objects within each distance range into three difficulty levels based on the percentile of the number of points.

The results show that detectors trained with \Ours{} consistently outperform those trained with OV-SCAN~\cite{ovscan} across all distances and difficulty levels. In particular, even in the \emph{Low} setting, where only a small number of points are observed and annotation is therefore challenging, \Ours{} shows improvement in mAP. This suggests that our annotations are more effective than OV-SCAN in training detectors to handle partially observed point clouds.

Furthermore, \Ours{} also improves performance across all image-based visibility levels defined by nuScenes~\cite{nuscenes}. We conjecture that this improvement comes from the different treatment of low-visibility instances. OV-SCAN~\cite{ovscan} relies on selective alignment and tends to filter out instances with small or low-quality masks, whereas \Ours{} uses a generation model~\cite{fastsam3d} that is more robust to mask quality to construct amodal 3D assets and derive 3D bounding boxes. As a result, \Ours{} can provide more reliable annotations even under severe partial visibility, leading to higher mAP.

\end{document}